\documentclass [conference]{IEEEtran}
\usepackage{epsfig,amsmath,amstext}
\usepackage{mymacros}
\usepackage{graphicx,latexsym}
\usepackage{sbsfigure}
\usepackage{color}
\usepackage{amssymb}

\graphicspath{{./figures/}}
\DeclareGraphicsExtensions{.pdf,.png}

\IEEEoverridecommandlockouts
\begin{document}

\title{Coded Federated Learning}

\author{\IEEEauthorblockN{Sagar Dhakal, Saurav Prakash, Yair Yona$^*$\thanks{$^*$Currently with Qualcomm Inc.}, Shilpa Talwar, Nageen Himayat}
\IEEEauthorblockA{Intel Labs\\
Santa Clara, CA 94022 \\
\{sagar.dhakal, saurav.prakash, shilpa.talwar, nageen.himayat\}@intel.com, $^*$yyona@qti.qualcomm.com}
\thanks{©2019 IEEE}}

\maketitle

\begin{abstract}
Federated learning is a method of training a global model from decentralized data distributed across client devices. Here, model parameters are computed locally by each client device and exchanged with a central server, which aggregates the local models for a global view, without requiring sharing of training data. The convergence performance of federated learning is severely impacted in heterogeneous computing platforms such as those at the wireless edge, where straggling computations and communication links can significantly limit timely model parameter updates. This paper develops a novel coded computing technique for federated learning to mitigate the impact of stragglers. In the proposed Coded Federated Learning (CFL) scheme, each client device privately generates parity training data and shares it with the central server only once at the start of the training phase. The central server can then preemptively perform redundant gradient computations on the composite parity data to compensate for the erased or delayed parameter updates. Our results show that CFL allows the global model to converge nearly four times faster when compared to an uncoded approach.
\end{abstract}


\section{Introduction}\label{intro}
Distributed machine learning (ML) over wireless networks has been gaining popularity recently, as training data (e.g. video images, health related measurements, traffic/crowd statistics, etc.) is typically located at wireless edge devices. Smartphones, wearables, smart vehicles, and other IoT devices are equipped with sensors and actuators that generate massive amounts of data.  Conventional approach to training ML model requires colocating the entire training dataset at the cloud servers. However,  transferring  client data from the edge to the cloud may not always be feasible. As most client devices have connectivity through wireless links, uploading of training data to the cloud may become prohibitive due to bandwidth limitations. Additionally, client data may be private in nature and  may not be shared with the cloud.

To overcome the challenges in cloud computing, federated learning (FL) has been recently proposed with the goal to leverage significant computation, communication and storage resources available at the edge of wireless network. For example, in \cite{mcMahan2017,bonawitz2017} federated learning is performed to improve the automatic text prediction capability of Gboard. Federated learning utilizes a federation of clients, coordinated by a central server, to train a global ML model such that client data is processed locally and only model parameters are exchanged across the network. One  key characteristic of federated learning is non-iid (independently and identically distributed) training data, where data stored locally on a device does not represent the population distribution \cite{zhao2018}. Further, devices generate highly disparate amount of training data based on individual usage pattern for different applications and services. Therefore, to train an unbiased global model, the central server needs to receive partial gradients from a diverse population of devices in each training epoch. However, due to the heterogeneity in compute and wireless communication resources of edge devices, partial gradient computations arrive in an unsynchronized and stochastic fashion.  Moreover, a small number of significantly slower devices and unreliable wireless links, referred to as straggler links and nodes, may  drastically prolong training time.

Recently, there is an emerging field of research, namely $coded$ $computing$, that applies concepts from error correction coding to mitigate straggler problems in distributed computing \cite{li2019,tandon2017,reisizadeh2019coded}. In \cite{li2019} polynomial coded regression has been proposed for training least-square models. Similarly, \cite{tandon2017} proposes a method named gradient coding wherein distributed computation and encoding of the gradient at worker clients are orchestrated by a central server. In \cite{reisizadeh2019coded} coding for distributed matrix multiplication is proposed, where an analytical method is developed to calculate near-optimal coding redundancy. However, the entire data needs to be encoded by the master device before assigning portions to compute devices. Coded computing methods based on centralized encoding, as proposed in \cite{li2019, tandon2017,reisizadeh2019coded,lee2018,karakus2019}, are not applicable in federated learning, where training data is located at different client nodes. Distributed update method using gossip algorithm has been proposed for learning from decentralized data, but it suffers from slow convergence due to lack of synchronization \cite{su2009}. Synchronous update method proposed in \cite{mcMahan2017} selects random worker nodes in each mini-batch, without considering compute heterogeneity, wireless link quality and battery life, which may cause the global model to become stale and diverge.

{\bf Our contribution}: In this paper, we develop a novel scheme named $Coded$ $Federated$ $Learning$ (CFL) for linear regression workloads. Specifically, we modify our coded computing solution in \cite{dhakal2018}, designed for a centrally available dataset, to develop a distributed coding algorithm for learning from decentralized datasets. Based on statistical knowledge of compute and communication heterogeneity, near-optimal coding redundancy is calculated by suitably adapting a two-step optimization framework given in \cite{reisizadeh2019coded}. While \cite{reisizadeh2019coded} requires the entire data to be centrally located for encoding, in our proposed CFL each device independently scales its local data and generates parity from the scaled local data to facilitate an unbiased estimate of the global model at the server. Only the parity data is sent to the central server while the raw training data and the generator matrix are kept private. During training each worker performs gradient computations on a subset of its local raw data (systematic dataset) and communicates the resulting partial gradient to the central server. The central server combines parity data received from all devices, and during each epoch computes partial gradient from the composite parity data. Due to these redundant gradient computations performed by the server,  only a subset of partial gradients from the systematic dataset  is required for reliably estimating the full gradient. This effectively removes the tail behavior, observed in uncoded FL, which is dominated by delays in receiving partial gradients from straggler nodes and links. Unlike \cite{reisizadeh2019coded}, the proposed CFL has no additional cost for decoding the partial gradients computed from the parity data.

This paper is organized as follows.  Section \ref{fed} outlines federated learning method for linear regression model. Section \ref{cgd} describes the coded federated learning algorithm. Numerical results are given in Section \ref{results} and Section \ref{con} ends with concluding remarks.

\section{Federated Learning} \label{fed}
We consider the scenario where training data is located at edge devices. In particular, the $i$-th device, $i=1,\dots,n$,  has $({\bf X}^{(i)},{\bf y}^{(i)})$ local database having $\ell_{i}\geq 0$ training data points given as
\begin{eqnarray*}
{\bf X}^{(i)} =
\left(
  \begin{array}{c}
    {\bf x}^{(i)}_1\\
    \vdots\\
    {\bf x}^{(i)}_{\ell_i}\\
  \end{array}
\right),
{\bf y}^{(i)} =
\left(
  \begin{array}{c}
    y^{(i)}_1\\
    \vdots\\
    y^{(i)}_{\ell_i}\\
  \end{array}
\right),
\end{eqnarray*}

\noindent where each training data-point ${\bf x}^{(i)}_k\in \mathbb{R}^{1\times d}$ is associated to a scalar label $y^{(i)}_k \in \mathbb{R}$. In a supervised machine learning problem, training is performed to learn the global model  $\boldsymbol\beta \in \mathbb{R}^d$ with $d$ as the fixed model size. Under a linear model assumption, the totality of training data points $m = \sum_{i=1}^n \ell_i$ can be represented as ${\bf y} = {\bf X \boldsymbol\beta} + {\bf z}$, where ${\bf X} = [{\bf X}^{(1)T}, \dots, {\bf X}^{(n)T}]^T$, ${\bf y} = [{\bf y}^{(1)T}, \dots, {\bf y}^{(n)T}]^T$, and ${\bf z} \in \mathbb{R}^{m\times 1}$ is measurement noise typically approximated as Gaussian iid samples. In gradient descent methods the unknown model is iteratively estimated by computing $\boldsymbol\beta^{(r)}$ at the $r$-th epoch, evaluating a gradient associated to the squared error cost function
\begin{eqnarray}
\lb{cost}
f(\boldsymbol\beta^{(r)}) = ||{\bf X} \boldsymbol\beta^{(r)} - \bf y||^2.
\end{eqnarray}

The gradient of the cost function in Eq.~(\ref{cost}) is given by
\begin{eqnarray}
\lb{grad}
\nabla_{\boldsymbol\beta}f({\boldsymbol\beta}^{(r)}) &=& {\bf X}^{T}({\bf X} {\boldsymbol\beta}^{(r)} - \bf y)\nonumber\\
&=& \sum_{i=1}^{n}\sum_{k=1}^{\ell_i}{\bf x}^{(i)T}_k({\bf x}^{(i)}_k {\boldsymbol\beta}^{(r)} - y^{(i)}_k).
\end{eqnarray}

Equation (\ref{grad}) decomposes the gradient computation into an inner sum of partial gradients that each device can locally compute and communicate to a central server, and an outer sum that a central server computes by aggregating the received partial gradients. Next $\boldsymbol\beta^{(r)}$  is updated by the central server according to
\begin{eqnarray}
\lb{updateBeta}
\boldsymbol\beta^{(r+1)} = \boldsymbol\beta^{(r)}-\frac{\mu}{m}\nabla_{\boldsymbol\beta}f({\boldsymbol\beta}^{(r)}),
\end{eqnarray}
where $\mu\geq0$ is an update parameter, and  $\boldsymbol\beta^{(0)}$ may be initialized arbitrarily. The paradigm of federated learning method is this recursive computation of inner sums happening at each device, followed by communication of the partial gradients to the central server; and the outer sum being computed by the central server to update of the global model, followed by communication of the updated global model to the edge devices. Equations (\ref{grad}) and (\ref{updateBeta}) are performed in tandem until sufficient convergence is achieved.

Performance of federated learning is limited by recurring delays in computing and communication of partial gradients.  A few straggler links and devices may keep the master device waiting for partial gradients in each training epoch. To capture this behavior, next we describe a simple model for computing and communication delays.

\subsection{Model for Computing and Communication Delays}\lb{delayModel}
In a heterogeneous computing platform like wireless edge computing, each device may have different processing rates, memory constraints, and active processes running on them. One approach to statistically represent the compute heterogeneity is to model the computation time for the $i$-th device by a shifted exponential random variable $T_{c_i}$ given as
\begin{eqnarray}
\lb{computeModel}
T_{c_i} = T_{c_{i,1}} + T_{c_{i,2}},
\end{eqnarray}

\noindent where $T_{c_{i,1}}= \ell_i a_i$ represents time to process $\ell_i$  training data points with each point requires $a_i$ seconds. $T_{c_{i,2}}$ is the stochastic component of the compute time that models randomness coming from memory read/write cycles during the Multiply-Accumulate (MAC) operations. The exponential probability density function (pdf) of $T_{c_{i,2}}$ is $p_{T_{c_{i,2}}}(t) = \gamma_i e^{-\gamma_i t},  t\geq 0$, where $\gamma_i =  \frac{\mu_i}{l_i}$, $\mu_i$ is memory access rate to read/write every training data (measured in per second unit).

The round-trip communication delay in each epoch includes the download time $T_{d_i}$ from the master node communicating an updated model to the $i$-th device, and the upload time $T_{u_i}$ from the $i$-th device communicating the partial gradient to the master node. The wireless communication links between the master device and worker nodes exhibit stochastic fluctuations in link quality. In order to maintain reliable communication service, it is a general practice to periodically measure link quality to adjust the achievable data rate. In particular, the wireless link between a master node and the $i$-th worker device can be modeled by a tuple $(r_i,p_i)$, where $r_i$  is the achievable data rate (in bits per second per Hz) with link erasure probability smaller than $p_i$ \cite{3gpp}. Therefore, the number of transmissions $N_i$ required before the first successful communication from the $i$-th device has a geometric distribution given by
\begin{eqnarray}
\lb{commModel}
Pr\{N_i=t\} = p^{t-1}_i(1-p_i), \,\, t = 1,2,3,...
\end{eqnarray}

\noindent It is a typical practice to dynamically adapt the data rate $r_i$ with respect to the changing quality of the wireless link while maintaining a constant erasure probability $p$ during the entire gradient computation. Further, without loss of generality,  we can assume uplink and downlink channel conditions are reciprocal. Downlink and uplink communication delays are random variables given by
\begin{eqnarray}
\lb{commTime}
T_{d_i} = N_i \tau_i,
\end{eqnarray}

\noindent where $\tau_i =\frac{x}{r_i W}$ is the time to upload (or download) a packet of size $x$ bits containing partial gradient (or model) and $W$ is the bandwidth in Hz assigned to the $i$th worker device. Therefore, the total time taken by the $i$-th device to receive the updated model, compute and successfully communicate the partial gradient to the master device is
\begin{eqnarray}
\lb{totDelay}
T_i = T_{c_i} + T_{d_i} + T_{u_i}.
\end{eqnarray}
It is straightforward to calculate the average delay given as
\begin{eqnarray}
\lb{avgDelay}
E[T_i] = \ell_i \bigg(a_i +\frac{1}{\mu_i}\bigg) + \frac{2 \tau_i}{1-p}.
\end{eqnarray}

\section{Coded Federated Learning}\label{cgd}
The coded federated learning approach, proposed in this section, enhances the training of a global model in an heterogeneous edge environment by privately offloading part of the computations from the clients to the server. Exploiting statistical knowledge of channel quality, available computing power at edge devices, and size of training data available at edge devices, we calculate (1) the amount of parity data to be generated at each edge device to share with the master server once at the beginning of the training, and (2) the amount of raw data ($systematic$ $data$) to be locally processed by each device for computing partial gradient during each training epoch. During each epoch the server computes gradients from the composite parity data to  compensate for partial gradients that fail to arrive on time due to communication delay or computing delay. Parity data is generated using random linear codes and random puncturing pattern whenever applicable. The client device does not share the generator matrix and puncturing matrix with the master server. The shared parity data cannot be used to decode the raw data, thereby protecting privacy. Another advantage of our coding scheme is that it does not require an explicit decoding step. Next, we describe the proposed CFL algorithm in detail.

\subsection{Encoding of the training data}\label{encode}
We propose to perform a random linear coding at each device, say the $i$-th device on its training data set  $({\bf X}^{(i)},{\bf y}^{(i)})$, having $\ell_i$ data elements.  In particular, a random generator matrix ${\bf G}_i$, with elements drawn independently from standard normal distribution (or, iid Bernoulli($\frac{1}{2}$) distribution), is applied on the $weighted$ local training data set to obtain a coded training data set  $({\bf \tilde X}^{(i)},{\bf \tilde y}^{(i)})$. In matrix notation we can write
\begin{eqnarray}
\lb{codedData}
{\bf \tilde X}^{(i)} = {\bf G}_i {\bf W}_i {\bf X}^{(i)},\,\,  {\bf \tilde y}^{(i)} = {\bf G}_i {\bf W}_i {\bf y}^{(i)}
\end{eqnarray}

The dimension of ${\bf G}_i$ is $c \times \ell_i$, where the row dimension $c$ denotes the amount of parity data to be generated at each device. We refer to $c$  as $coding$ $redundancy$ and its  derivation is described in next section. Typically $c<<\sum_{i=1}^n \ell_i$. The matrix ${\bf W}_i$  is  $\ell_i \times \ell_i$  diagonal matrix that weighs each training data point. The weight matrix derivation is also deferred until the next section. It is to be noted that the locally coded training data set $({\bf \tilde X}^{(i)},{\bf \tilde y}^{(i)})$ is transmitted to the central server, while ${\bf G}_i$ and ${\bf W}_i$ are kept private.  At the central server, the parity data received from all client devices are combined to obtain the composite parity data set ${\bf \tilde X} \in \mathbb{R}^{c \times d}$ , and composite parity label ${\bf \tilde y}\in \mathbb{R}^{c \times 1}$ given as
\begin{eqnarray}
\lb{parityData}
{\bf \tilde X} = \sum_{i=1}^n {\bf \tilde X}^{(i)},\,\,  {\bf \tilde y} = \sum_{i=1}^n {\bf \tilde y}^{(i)}
\end{eqnarray}

\noindent Using Eqs.~(\ref{codedData}) and (\ref{parityData}) we can write
\begin{eqnarray}
\lb{parityX}
{\bf \tilde X} = \sum_{i=1}^n {\bf G}_i {\bf W}_i {\bf X}^{(i)} = {\bf G W X}
\end{eqnarray}

\noindent where ${\bf G} = [{\bf G}_1, \dots, {\bf G}_n]$ and ${\bf W}$ is a block-diagonal matrix given by
\begin{eqnarray*}
{\bf W} =
\left(
  \begin{array}{ccc}
    {\bf W}_1 & \dots & 0 \\
    \vdots &  & \vdots \\
    0 & \dots & {\bf W}_n \\
  \end{array}
\right).
\end{eqnarray*}

\noindent Similarly, we can write
\begin{eqnarray}
\lb{parityY}
{\bf \tilde y}  = {\bf G W y}
\end{eqnarray}

Equations.~(\ref{parityX}) and (\ref{parityY}) represent the encoding over the entire decentralized data set $({\bf X , y})$, performed implicitly in a distributed manner across host devices. Further, it is to be noted that $ {\bf G, W, X , y}$ are all unknown at the central server, thereby preserving privacy of raw training data of each  device.

\subsection{Calculation of coding redundancy} \label{red}
Let the $i$-th client device calculate partial gradient from $\tilde{\ell}_i$ local data points, and let $R_i (t;\tilde{\ell}_i)$  be an indicator metric representing the event that partial gradient computed and communicated by the $i$-th device is received at the master device within time $t$ measured from the beginning of each epoch. More specifically, $R_i (t;\tilde{\ell}_i)=\tilde{\ell}_i \textbf{1}_{\{T_i\leq t\}}$. Clearly, the return metric is either 0, or  $\tilde{\ell}_i$ . Next, we can define aggregate return metric as follows:
\begin{eqnarray}
\lb{aggReturn}
R(t;\tilde{\boldsymbol \ell}) = \sum_{i=1}^{n+1} R_i (t;\tilde{\ell}_i)
\end{eqnarray}

It is important to note in Eq.~(\ref{aggReturn}) that $(n+1)$-th  device represents the central server, and $\tilde{\ell}_{n+1}$  represents the number of parity data to be shared to the central server by each device. Next we find a load distribution policy ${\boldsymbol {\ell}^*}$ that provides an expected value of aggregate return equal to $m$  for a minimum waiting time $t^*$ in each epoch. Note that $m$ is the totality of  raw data points spread across $n$ edge devices.

For the computing and communication delay model given in Section~(\ref{delayModel}), we numerically found that the expected value of return metric from each device is a concave function of number of training data points processed at that device as shown in  Fig.~(\ref{fig:concave}) below.
  \begin{figure}[htb!]
  \centering
  \includegraphics[scale=0.4]{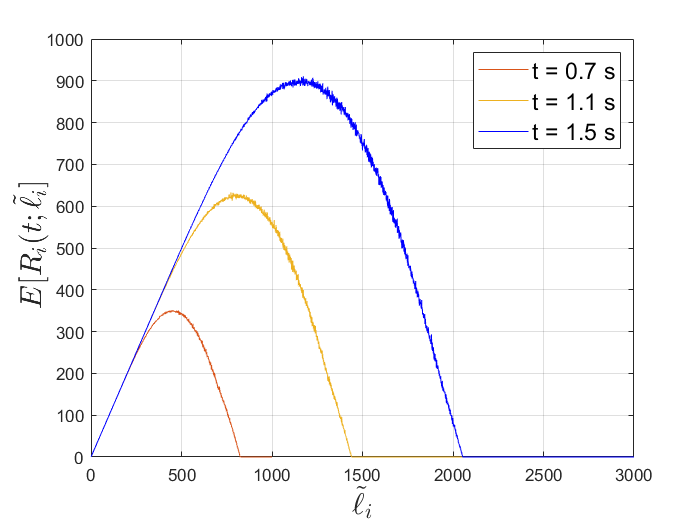}
  \caption{Expected value of individual return for different load assignments.}\label{fig:concave}
  \end{figure}

The probability that a device returns the partial gradient within a fixed epoch time, say $t = 0.7$ s in Fig.~(\ref{fig:concave}),  depends on number of raw data points used by that device. Intuitively, if number of raw data points $\tilde{\ell}_i$ is small, average computing and communication delays are small, therefore, the probability of return is larger. However, the expected return $E[R_i (t;\tilde{\ell}_i)]$ is small as well since this is bounded by $\tilde{\ell}_i$. Therefore, we observe that the expected return $E[R_i (t;\tilde{\ell}_i)]$ grows linearly with $\tilde{\ell}_i$ for small values of $\tilde{\ell}_i$. As the number of raw data points increases, the computing and communication delays increase resulting in decrease in the probability of return, and the expected return $E[R_i (t;\tilde{\ell}_i)]$ grows sub-linearly. Further increase in the number of raw data points will further decrease the probability of return to a point that the return time becomes, almost surely, larger than the epoch time $t= 0.7$ s, at which point the expected return $E[R_i (t;\tilde{\ell}_i)]$ becomes 0. It should also be noted that increasing the epoch time window to, say $t = 1.1$ s or $t = 1.5$ s, will allow a device extra time to process additional raw data while maintaining a larger probability of return. But the overall behavior stays the same. This clearly shows that there is an optimal number of training data points $\ell_i^* (t)$ to be evaluated at the $i$-th node that maximizes its average return for time $t$. More precisely, for a given time of return $t$,
\begin{eqnarray}
\lb{optiLk}
\ell_i^* (t) = \text{argmax}_{0 \leq \tilde{\ell}_i \leq {\ell}_i}E[R_i (t;\tilde{\ell}_i)]
\end{eqnarray}

Similarly, the optimal number of parity data to be processed at the central server for a given return time $t$ is
\begin{eqnarray}
\lb{optiLkMEC}
\ell_{n+1}^* (t) = \text{argmax}_{0 \leq \tilde{\ell}_{n+1} \leq c^{\text{up}}} E[R_{n+1} (t;\tilde{\ell}_{n+1})]
\end{eqnarray}

\noindent where $c^{\text{up}}$ denotes the maximum data the central server can receive from edge devices to limit the data transfer overhead. Next, from Eq.~(\ref{aggReturn}) we can note that maximum expected aggregate return is achieved by  maximizing expected return from each device separately. Finally, the optimal epoch time $t^*$ that makes the expected aggregate return equal to be $m$ is
\begin{eqnarray}
\lb{epochTime}
t^* = \text{argmin}_{t \geq 0}: m \leq E[R(t;{\boldsymbol \ell}^*(t))]\leq m +\epsilon,
\end{eqnarray}

\noindent where $\epsilon \geq 0$ is a tolerance parameter. The coding redundancy $c$, which is the row dimension of generator matrix ${\bf G}_i$, is given by  $c=\ell_{n+1}^* (t^*)$ and the number of raw data points to be processed at $i=1,\dots,n$ devices are $\ell_{i}^* (t^*)$.

\subsection{Weight matrix computation} \label{weight}

The $i$-th device uses the weight matrix ${\bf W}_i$  which is a $\ell_i \times \ell_i$  diagonal matrix. The diagonal coefficients of the weight matrix corresponding to $k =1,\dots,\ell_{i}^* (t^*)$ data points are given by
\begin{eqnarray}
\lb{probNoReturn}
w_{ik} = \sqrt{Pr\{T_i \geq t^{*}\}}.
\end{eqnarray}

Thus, weights are calculated from the probability that the central server does not receive partial gradient from the $i$-th device within epoch time $t^*$. For a given load partition $\ell_{i}^* (t^*)$,  this probability can be directly computed by the $i$-th edge device using probability distribution function of computation times and communication link delays. Further, it should be noted that there are $(\ell_i - \ell_{i}^* (t^*))$ uncoded data points that are punctured and never processed at the $i$-th edge device. The diagonal coefficients of the weight matrix corresponding to the punctured data points are set as $w_{ik}= 1$. Puncturing of raw data provides another layer of privacy as each device can independently select which data points to puncture.

\subsection{Aggregation of partial gradients}
In each epoch we have two types of partial gradients available at the central server. The central server computes the normalized aggregate of partial gradients from the composite parity data $({\bf \tilde X},{\bf \tilde y})$ as
\begin{eqnarray}
\lb{codedGrad}
\frac{1}{c}{\bf \tilde X}^{T}({\bf \tilde X} {\boldsymbol\beta}^{(r)} - {\bf \tilde y})
&=&{\bf X}^T {\bf W}^T\biggl(\frac{1}{c}{\bf G}^T{\bf G}\biggr){\bf W}({\bf X}{\boldsymbol\beta}^{(r)}-{\bf y})\nonumber\\
&\approx& {\bf X}^T {\bf W}^T{\bf W}({\bf X}{\boldsymbol\beta}^{(r)}-{\bf y})\nonumber\\
\!\!\!\!\!\!&=&\!\!\!\!\!\! \sum_{i=1}^{n}\sum_{k=1}^{\ell_i}\!\!\! w_{ik}^2 {\bf x}^{(i)T}_k({\bf x}^{(i)}_k {\boldsymbol\beta}^{(r)} - y^{(i)}_k)
\end{eqnarray}

\noindent In deriving above identity we have applied the weak law of large numbers to replae the quantity $\frac{1}{c}{\bf G}^T{\bf G}$ by an identity matrix for sufficiently large value of $c$. The other set of partial gradients are computed by edge devices on their local uncoded data and transmitted to the master node. The master node waits for the partial gradients only until optimized epoch time $t^*$ and aggregates them.  The expected value of sum of  partial gradients received from edge devices by time  $t^*$ is given by
\begin{eqnarray}
\lb{receivedGrad}
&&\sum_{i=1}^{n}\sum_{k=1}^{\ell_i}{\bf x}^{(i)T}_k({\bf x}^{(i)}_k {\boldsymbol\beta}^{(r)} - y^{(i)}_k) Pr\{T_i \leq t^*\}\nonumber\\
&& = \sum_{i=1}^{n}\sum_{k=1}^{\ell_i}{\bf x}^{(i)T}_k({\bf x}^{(i)}_k {\boldsymbol\beta}^{(r)} - y^{(i)}_k) (1-w_{ik}^2),
\end{eqnarray}

\noindent where we have used Eq.~(\ref{probNoReturn}) to replace the probability of return. The master can simply combine two sets of gradients from Eqs.~(\ref{codedGrad}) and (\ref{receivedGrad}) to obtain $\nabla_{\boldsymbol\beta}f({\boldsymbol\beta}^{(r)})$, which approximately represents gradient over entire data as given by Eq.~(\ref{grad}).

\section{Numerical Results}\label{results}
We consider a wireless network comprising one master node and  24 edge devices. Training data ${\bf X, y}$ is generated from
${\bf y} = {\bf X \boldsymbol\beta} + {\bf n}$,
where each element $X_{kj}$ have iid Normal distribution, measurement noise is AWGN, and signal to noise ratio (SNR) is 0 dB.  Dimension of model ${\bf \boldsymbol\beta}$  is set to $d = 500$. Each edge device has $\ell_i = 300$ training data points for $i = 1,\dots,24$. Learning rate $\mu$ is set to $0.0085$.

To model heterogeneity across devices, we define a compute heterogeneity factor $0\leq\nu_{\text{comp}}<1$. We generate 24 MAC rates given by $\text{MACR}_i=(1-\nu_{\text{comp}})^i \times$ 1536 KMAC per second, for $i = 0, \dots, 23$, and randomly assign a unique value to each edge device. Note that compute heterogeneity is more severe for larger values of $\nu_{\text{comp}}$. As each training point requires $d = 500$ MAC operations, the deterministic component of computation time per training data at the $i$-th device (as defined in Section~\ref{delayModel})  is $a_i = \frac{d}{\text{MACR}_i}$. We assign a 50 \% memory access overhead per training data point as $\mu_i = \frac{2}{a_i}$. Lastly, the compute rate at master node is assumed to be 10 times faster than the fastest edge device, i.e.,  the MAC rate of the master node is set as 15360 KMAC per second.

 Similarly,  we define a link heterogeneity factor $0\leq\nu_{\text{link}}<1$. We generate 24 link throughput given by $(1-\nu_{\text{link}})^i \times$ 216  Kbits per second, for $k = 0, \dots, 23$ and randomly assign a unique value to each link. Note that link heterogeneity is more severe for larger values of  $\nu_{\text{link}}$. Each communication packet is a real-valued vector of size $d$, where each element in the vector is represented by 32 bit floating point. Packet size is calculated accordingly with additional 10\% overhead for header. The link failure rate is set as $p_i=0.1$ for all links.

\begin{figure}[htb!]
\centering
\includegraphics[scale=0.4]{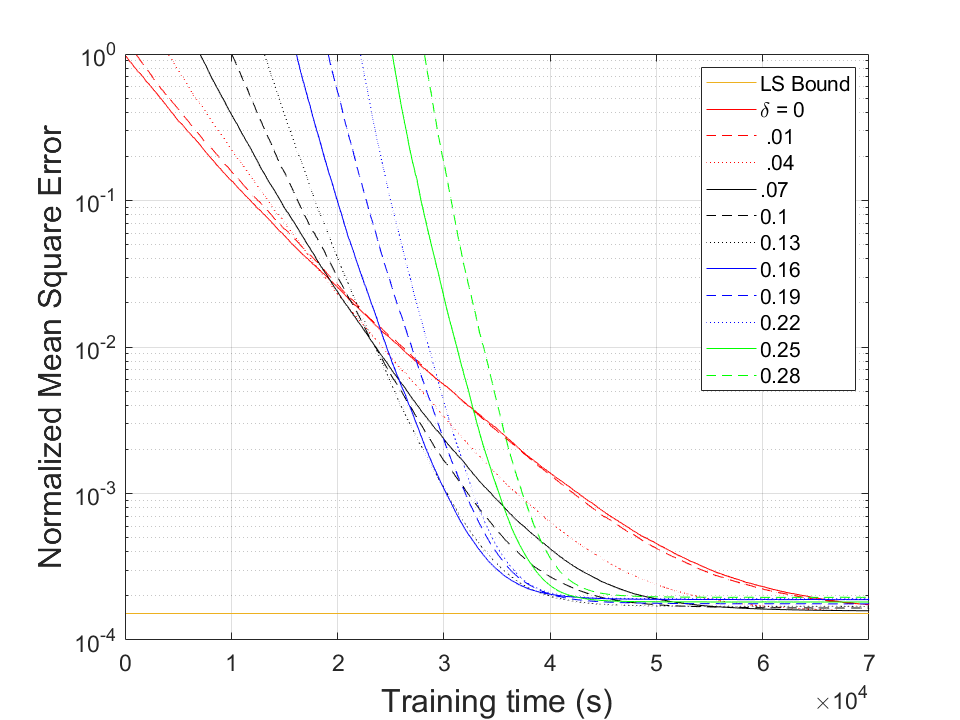}
\caption{Convergence time of CFL for different coding redundancy values.}\label{fig:converge}
\end{figure}

\begin{figure}[htb!]
\centering
\includegraphics[width=0.9\linewidth]{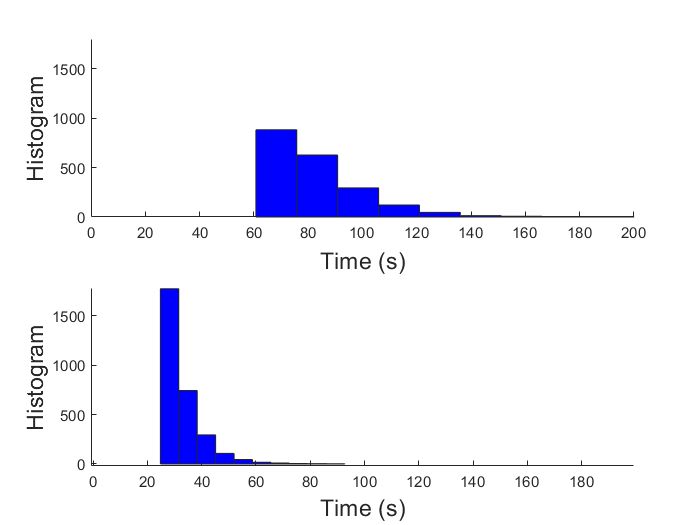}
\caption{Time to receive $m$ partial gradients in uncoded federated learning (top), and $m-c$ partial gradients in coded federated learning (bottom).}\label{fig:epochTimeCodedUncoded}
\end{figure}

In Fig.~(\ref{fig:converge}) we show the convergence of gradient descent algorithm in terms of normalized mean square error (NMSE) as a function of training time. NMSE in the $r$-th epoch is defined as $\frac{||{\boldsymbol\beta}^{(r)} - \boldsymbol\beta||^2}{||\boldsymbol\beta||^2}$. The degree of heterogeneity  is set at at $\nu_{\text{comp}}= 0.2$ and $\nu_{\text{link}}=0.2$. The performance is benchmarked against the least square (LS) bound. In order to quantify the coding redundancy, we have introduce a redundancy metric $\delta = \frac{c}{\sum_i^n \ell_i}$. Clearly,  $\delta=0$ represents uncoded federated learning, which exhibit a slow convergence rate due to straggler effect. In Fig.~(\ref{fig:epochTimeCodedUncoded}) the top plot shows the histogram of time to receive $m$ partial gradients in uncoded federated learning, which  exhibits a tail extending beyond 150 s.

As $\delta$  is increased from 0 to 0.28, the convergence rate of CFL increases. A larger value of $\delta$ provides more parity data enabling the master node to disregard straggling links and nodes. The bottom plot of Fig.~(\ref{fig:epochTimeCodedUncoded}) shows the histogram of time to receive $m -c$ partial gradients in CFL with $\delta=0.13$. By comparison of the top and bottom plots, the long tail observed in uncoded FL can be attributed to the last $c$ partial gradients.  By performing an in-house computation of  partial gradients form $c$ parity data points,  the master node receives, on an average $\sum_i^n \ell_i -c$ partial gradients, in a much smaller epoch time $t^*$. But, a large value of $\delta$ will also lead to a large communication cost to transfer parity data to the master node, which delays the start of training. Therefore, an arbitrarily chosen $\delta$ may, at times, lead to a worse convergence behavior than the uncoded solution. From Fig.~(\ref{fig:converge}) we can observe the effect of coding in creating initial delays for different values of $\delta$. Clearly, it is more prudent to select a particular coded solution based on the required accuracy of the model to be estimated. For example, at an NMSE of  $0.1$ the uncoded learning outperforms all coded solutions, whereas at an NMSE of  $10^{-3}$, coded solution with $\delta=0.16$ provides the minimum convergence time.

\begin{figure}[htb!]
\centering
\includegraphics[scale=0.4]{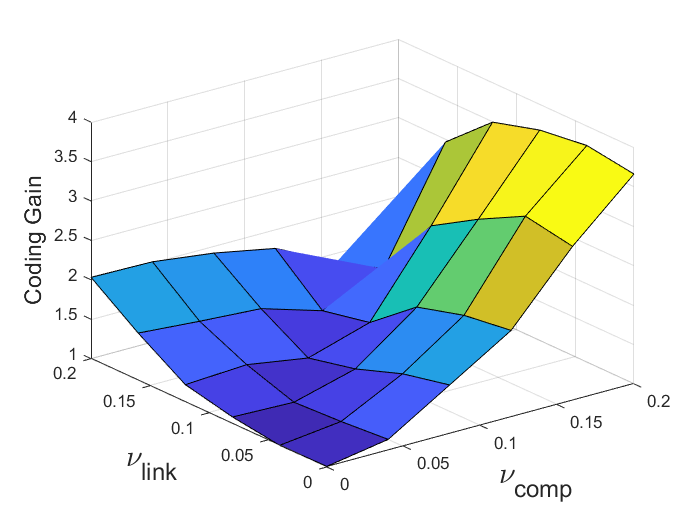}
\caption{Coding gain at different heterogeneity values.}\label{fig:surface}
\end{figure}

In Fig.~(\ref{fig:surface}) we plot the ratio of  convergence times, referred hereforth as $coding$  $gain$,  for optimal coded learning to that of uncoded learning for different heterogeneity values. Here convergence time is measured as time to achieve an  NMSE $\leq 3 \times 10^{-4}$. The coding gain measures how fast the coded solution converges to a required NMSE compared to the uncoded method. As can be observed, depending on  heterogeneity level defined by the tuple $(\nu_{\text{comp}},\nu_{\text{link}})$, the CFL provides between 1 to nearly 4 times coding gain over uncoded FL. At the maximum heterogeneity of  $(0.2,0.2)$, maximum  coding gain is achieved. Whereas at heterogeneity of  $(0,0)$ (a homogeneous scenario), the coding gain approaches unity.

\begin{figure}[htb!]
\centering
\includegraphics[scale=0.4]{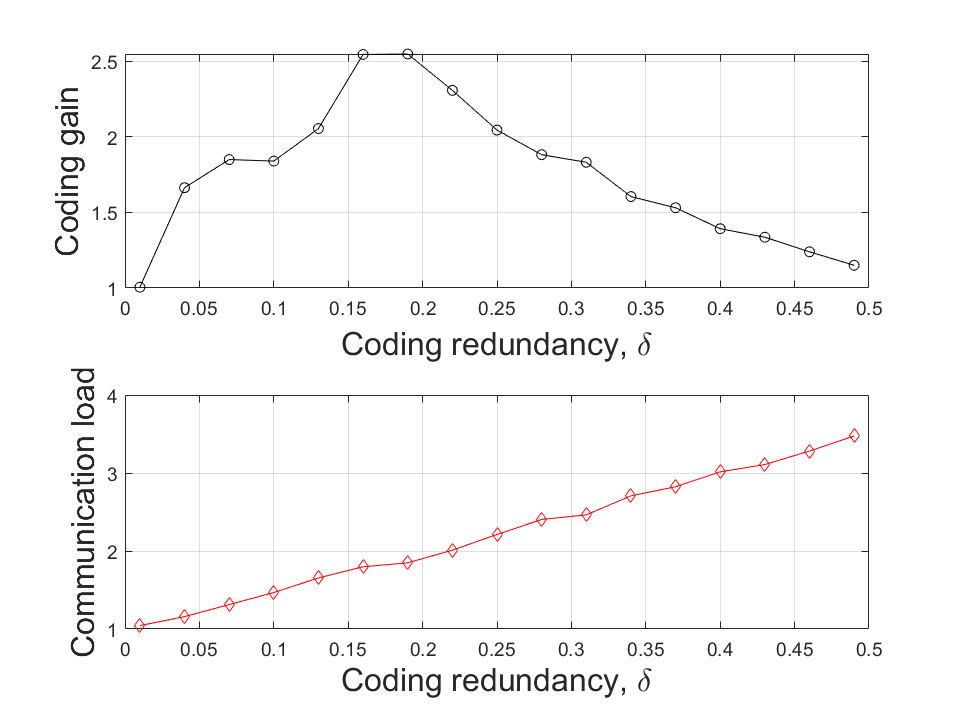}
\caption{Coding gain against communication load for $\nu_{\text{comp}}=0.4,\nu_{\text{link}}=0.4$.}\label{fig:gainVsBwP4}
\end{figure}

In Fig.~(\ref{fig:gainVsBwP4}) the top plot shows coding gain for different values of coding redundancy metric $\delta$, and the bottom figure shows corresponding increase in communication load for parity data transmission.  For a target NMSE of $1.8 \times10^{-4}$, when heterogeneity is $\nu_{\text{comp}}= 0.4$ and $\nu_{\text{link}}=0.4$, CFL converges 2.5 times faster than uncoded FL at $\delta = 0.16$  while incurring 1.8 times more data bits to be transferred. Similarly,  we observed that when the heterogeneity is set at $\nu_{\text{comp}}= 0.2$ and $\nu_{\text{link}}=0.2$, maximum coding gain of 1.6 could be achieved at $\delta = 0.13$ for an associated cost of transmitting 1.6 times more data compared to uncoded FL. 

\section{Conclusion}\label{con}
We have developed a novel coded computing methodology targeting linear distributed machine learning (ML) from decentralized training data sets for mobile edge computing platforms. Our coded federated learning method utilizes statistical knowledge of compute and communication delays to independently generate parity data at each device. We have introduced the concept of probabilistic weighing of the parity data at each device to remove bias in gradient computation as well as to provide an additional layer of data privacy. The parity data shared by each device is combined by the central server to create a composite parity data set, thereby achieving distributed coding across decentralized data sets. The parity data allows the central server to perform gradient computations that substitute or replace late-arriving or missing gradients from straggling client devices, thus clipping the tail behavior during synchronous model aggregation at each time epoch. Our results show that the coded solution results in nearly four times faster convergence compared to uncoded learning. Furthermore, the raw training data, and the generator matrices are always kept private at each device, and there is no decoding of partial gradients required at the central server.

To the best of our knowledge, this is the first paper that develops a coded computing scheme for federated learning of linear models. Our approach gives a flexible framework for dynamically tuning the tradeoff between coding gain and demand on channel bandwidth, based on the targeted accuracy of the model. Many future directions are suggested by early results in this paper. One important extension is to develop solutions for non-linear ML workloads using non-iid data and client selection. Another key direction would be to analyze privacy guarantees of coded federated learning. For instance, in \cite{zhou2009} authors have formally shown privacy resulting from a simple random linear transformation of raw data.
{\refsize
\bibliography{bibl}
\bibliographystyle{ieeetr}
}

\end{document}